\documentclass[sigconf]{acmart}
\copyrightyear{2025}
\acmYear{2025}
\setcopyright{rightsretained}
\acmConference[MM '25] {Proceedings of the 33rd ACM International Conference on Multimedia}{October 27--31, 2025}{Dublin, Ireland.}
\acmBooktitle{Proceedings of the 33rd ACM International Conference on Multimedia (MM '25), October 27--31, 2025, Dublin, Ireland}
\acmISBN{979-8-4007-2035-2/2025/10}
\acmDOI{10.1145/3746027.3761989}
\AtBeginDocument{%
  }

\usepackage{pifont}

\usepackage{multirow}
\begin{document}

%%
%% The "title" command has an optional parameter,
%% allowing the author to define a "short title" to be used in page headers.
\title{Identity-Preserving Text-to-Video Generation via Training-Free Prompt, Image, and Guidance Enhancement
% Training-free Enhancement for Identity-Preserving Text-to-Video Generation 
}

%%
%% The "author" command and its associated commands are used to define
%% the authors and their affiliations.
%% Of note is the shared affiliation of the first two authors, and the
%% "authornote" and "authornotemark" commands
%% used to denote shared contribution to the research.
\author{Jiayi Gao}
\authornote{Both authors contributed equally to this research.}
\email{gaojiayi0728@gmail.com }
\orcid{1234-5678-9012}
% \author{Jiayi Gao}
% \authornotemark[1]
% \email{gaojiayi0728@gmail.com}
\affiliation{%
  \institution{Wangxuan Institute of Computer Technology, Peking University, }
  \city{Beijing}
  \country{China}
}

% \author{
% Changcheng Hua}
% \authornote{Both authors contributed equally to this research.}
% \email{hcc@stu.pku.edu.cn}
% \orcid{1234-5678-9012}
\author{Changcheng Hua}
\authornotemark[1]
\email{hcc@stu.pku.edu.cn}
\affiliation{%
  \institution{Wangxuan Institute of Computer Technology,Peking University}
  \city{ Beijing}
  \country{China}
}

\author{Qingchao Chen}
\email{qingchao.chen@pku.edu.cn}
\affiliation{%
  \institution{National Institute of Health Data Science,Peking University}
  \city{Beijing}
  \country{China}
}

\author{Yuxin Peng}
\email{pengyuxin@pku.edu.cn}
\affiliation{%
  \institution{Wangxuan Institute of Computer Technology, Peking University}
  \city{Beijing}
  \country{China}
}

\author{Yang Liu}
\authornote{Corresponding author.}
\email{yangliu@pku.edu.cn}
\affiliation{%
  \institution{Wangxuan Institute of Computer Technology, Peking University}
  \city{Beijing}
  \country{China}
}

\begin{abstract}
Identity-preserving text-to-video (IPT2V) generation creates videos faithful to both a reference subject image and a text prompt. While fine-tuning large pretrained video diffusion models on ID-matched data achieves state-of-the-art results on IPT2V, data scarcity and high tuning costs hinder broader improvement.
We thus introduce a \textit{\textbf{T}raining-Free \textbf{P}rompt, \textbf{I}mage, and \textbf{G}uidance \textbf{E}nhancement}(\textbf{TPIGE}) framework that bridges the semantic gap between the video description and the reference image and design sampling guidance that enhances identity preservation and video quality, achieving performance gains at minimal cost.
Specifically, we first propose \textit{\ding{192}Face Aware Prompt Enhancement}, using GPT-4o to enhance the text prompt with facial details derived from the reference image. We then propose \textit{\ding{193}Prompt Aware Reference Image Enhancement},  leveraging an identity-preserving image generator to refine the reference image, rectifying conflicts with the text prompt. The above mutual refinement significantly improves input quality before video generation. Finally, we propose \textit{\ding{194}ID-Aware Spatiotemporal Guidance Enhancement}, utilizing unified gradients to optimize identity preservation and video quality jointly during generation.
Our method outperforms prior work and is validated by automatic and human evaluations on a 1000-video test set—winning first place in the ACM Multimedia 2025 Identity-Preserving Video Generation Challenge, demonstrating state-of-the-art performance and strong generality. The code is available at https://github.com/Andyplus1/IPT2V.git. 
\end{abstract}

\ccsdesc[500]{Computing methodologies~Artificial intelligence}

%%
%% Keywords. The author(s) should pick words that accurately describe
%% the work being presented. Separate the keywords with commas.
\keywords{Identity-Preserving Video Generation; Prompt Learning; Identity Preserved Guidance;}
%% A "teaser" image appears between the author and affiliation
%% information and the body of the document, and typically spans the
%% page.

% \received{20 February 2007}
% \received[revised]{12 March 2009}
% \received[accepted]{5 June 2009}

%%
%% This command processes the author and affiliation and title
%% information and builds the first part of the formatted document.
\maketitle
\begin{figure*}
  \centering
  \includegraphics[width=0.7\textwidth]{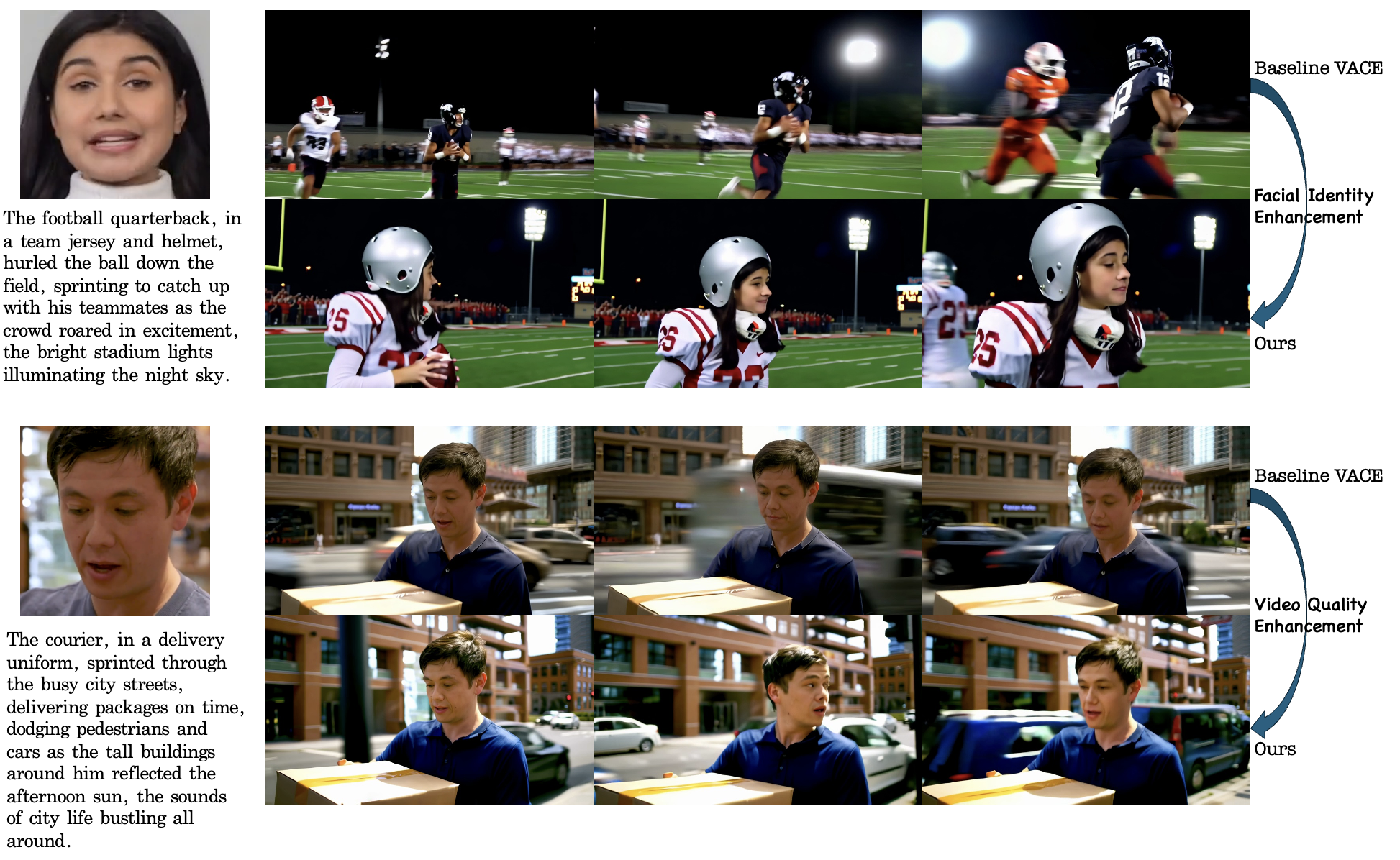}
  \caption{Our method produces higher-quality identity-preserving videos than baselines: faces are clearer and match the reference, while video quality better (motion is more natural, and the person is no longer static relative to the house)}
  
  \label{fig:teaser}
\end{figure*}
\section{Introduction}
Identity-preserving text-to-video generation \cite{dreamvideo, customvideo} takes a reference image and a text prompt as inputs and outputs a video that matches both.
Existing approaches fall into two camps: per-ID inference-time fine-tuning (PIFT)  \cite{motionbooth, Still-moving, magic-me} and inference-time tuning-free (ITF) \cite{ID-Animator,yuan2025identity,liu2025phantom,jiang2025vace}. PIFT's per-identity inference-time model adaptation is time-consuming and shows limited scalability. ITF methods employ a dedicated identity module that requires offline fine-tuning using paired identity video datasets. Although these methods achieve better quality and zero-shot deployment, their massive data and computation demands make continued optimization costly. 
To avoid such fine-tuning burden, we devise a completely \textit{training-free} strategy based ITF backbone for identity-preserving text-to-video generation that shows better performance.

Existing methods for identity-preserving video generation struggle to satisfy two objectives simultaneously: faithfully retaining the reference face and generating high quality video. As Figure \ref{fig:teaser}  shows, a model may either fail to insert the person at all (face is invisible because of the helmet) or insert the correct person while video quality degrades (the near-static figure and background building fails to match description `sprint').
A key reason is the semantic gap between the reference ID image and the text prompt, forcing the generator to trade off between identity preservation and prompt fidelity. This gap primarily arises from semantic conflicts between inputs: prompts describing face-obscuring roles
(e.g. helmeted football players in Figure \ref{fig:teaser} line 1) or appearance-altering attributes (e.g., gender/accessories, like the ``delivery package" in Figure \ref{fig:teaser} line 2) contradict the reference image, challenging the model's reconciliation capability.
Furthermore, current sampling strategy's guidance mainly focus on text condition, hindering joint optimization of identity preservation and perceptual video quality.

Thus we propose a \textit{\textbf{T}raining-Free \textbf{P}rompt, \textbf{I}mage, and \textbf{G}uidance \textbf{E}nhancement}(\textbf{TPIGE}) framework for the task, comprises three key components: \textit{\ding{192} Face Aware \textbf{P}rompt \textbf{E}nhancement}(\textbf{PE}), 
\textit{\ding{193} Prompt Aware Reference \textbf{I}mage \textbf{E}nhancement}(\textbf{IE}) and \textit{\ding{194} ID-Aware Spatiotemporal \textbf{G}uidance \textbf{E}nhancement}(\textbf{GE}). 
Specifically, \textit{\textit{\ding{192}Face Aware Prompt Enhancement} ensures text prompts are ID-aware: it uses GPT-4o \cite{hurst2024gpt} to automatically add detailed facial descriptions to original prompts (e.g., turning "The football quarterback" into "The football quarterback who is a person in her 20s with long black hair...") to assist facial generation. As shown in Figure. \ref{fig:teaser}, the generator thus better focus on these facial details.} \textit{\ding{193}Prompt Aware Reference Image Enhancement} uses an identity-preserving image generator to enhance reference images: it implants prompt-aligned ID attributes (e.g., occupational attire, facial expressions—such as regenerating a man with a delivery package, as in Figure. \ref{fig:teaser}), which visually reduces semantic conflict between identity images and text prompts. 
\textit{\ding{194} ID-Aware Spatiotemporal Guidance Enhancement} steers denoising by using the noise difference between a reference-conditioned strong model and a decaying reference-free weak model, jointly optimizing identity fidelity and video quality to meet broader evaluation criteria.

Our method achieved substantial improvements in identity preservation and video quality metrics. In addition, we introduce a tailored Mixture-of-Experts (MoE) strategy that selects and integrates the best-performing results from videos produced by different generation methods. On a 1,000-sample evaluation set, our method achieved the highest overall score across all metrics and won first place in the ACM MM 2025 Identity-Preserving Video Generation Challenge, providing strong evidence of its effectiveness.

The contributions of this paper are summarized as follows: 
\begin{itemize}
  \item We introduce the first training-free identity-preserving text-to-video generation framework, eliminating both inference-time per-ID tuning and costly post-training while retaining state-of-the-art performance.
  \item We propose Face Aware
Prompt Enhancement and Prompt Aware Reference Image Enhancement to bridge the semantic gap of the generation condition, and introduce  ID-Aware Spatiotemporal Guidance Enhancement to jointly optimize identity preservation and video quality during sampling.
  \item  
  Our \textbf{TPIGE}, equipped with an MoE strategy, won first place in the ACM MM 2025 Identity-Preserving Video Generation Challenge, achieving top performance on identity preservation and video quality metrics, as well as in a user study involving 3,000 1v1 comparison pairs.
\end{itemize}

\begin{figure*}[t]          % 使用 figure* 并放置在页面顶部
  \centering
  \includegraphics[width=0.85\textwidth]{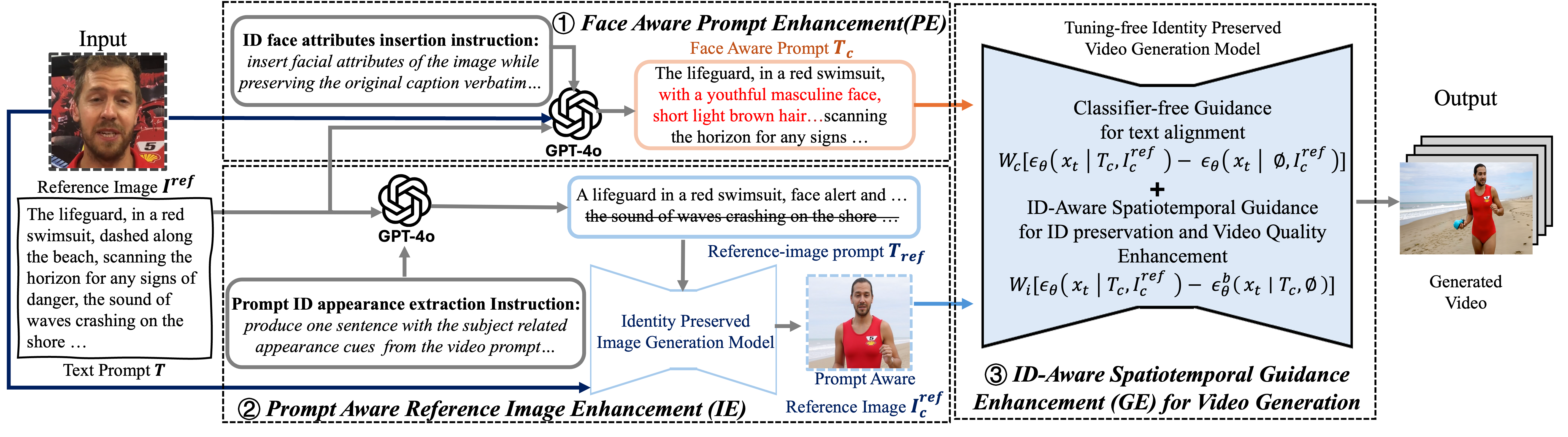}
  \caption{\textit{Pipeline of TPIGE}. TPIGE consists of three parts: (1) Face Aware
Prompt Enhancement enriches the prompt with facial attributes; (2) Prompt Aware Reference Image
Enhancement edits the reference image to incorporate prompt-aligned appearance cues; and (3) ID-Aware Spatiotemporal Guidance Enhancement guide the sampling for improved identity preservation and video quality.} 
  \label{pipeline}
\end{figure*}

\section{Related Work}

Diffusion models \cite{esser2024scaling, yang2024cogvideox, kong2024hunyuanvideo, gao2025conmo} have propelled significant progress in many downstream tasks \cite{magictime, controlnet, InstaDrag, evagaussians, cycle3d, ViewCrafter, li2025balancing,li2024diff} including identity-preserving generation \cite{dreamvideo, customvideo,xu2024semantic,luo2024generative,zhu2025customized}. Early approaches primarily relied on per-ID fine-tuning methods, such as MotionBooth \cite{motionbooth} and DreamVideo \cite{dreamvideo}, which incorporated reference content by fine-tuning model parameters or introducing additional modules. However, these methods required retraining for each new identity, greatly limiting scalability and practical deployment. To address these challenges, tuning-free strategies emerged, ACE++ \cite{mao2025ace++} and PhotoMaker \cite{photomaker} developed subject-preserved image generation models based on this approach. 
More recently, advanced models like Phantom \cite{liu2025phantom} and VACE \cite{jiang2025vace} have demonstrated the capability to generate consistent multi-subject videos in open-domain scenarios \cite{liang2025movie, chen2025multi}, steadily closing the performance gap with commercial solutions like Hailuo \cite{hailuo2024} and Vidu \cite{Vidu}. Nevertheless, these methods require collecting large amounts of data and time-consuming post-training. In contrast, our approach builds on existing open-source tuning-free models and produces higher-quality videos through enhancement strategies, without the need for individual fine-tuning.

\section{Methodology}

\subsection{Overview}
As shown in Figure \ref{pipeline}, our proposed TPIGE enhances the identity-preserving video generation  model \cite{jiang2025vace} through three improvements. Given the input condition: text prompt $T$ and reference human face ID image $I^{ref}$, to close the semantic gap between reference image and text prompt, we design two steps: \textbf{Face Aware Prompt Enhancement} and \textbf{Prompt Aware Reference Image Enhancement}. First, GPT-4o \cite{hurst2024gpt} is leveraged to automatically add fine facial descriptors to the text prompt at this stage, resulting in an enhanced prompt $T_c$ with explicit identity cues. Then identity-preserving image generator edits the reference face image to embed prompt-specific attributes (e.g., uniform, expression), yielding a less-conflict portrait $I^{ref}_c$ in the stage. $T_c$ and $I^{ref}_c$ are the input conditions for the video generation model \cite{jiang2025vace}. During generation, we apply \textbf{ID-Aware Spatiotemporal Guidance 
Enhancement}, ensuring gradients are directed toward the target distribution to generate videos with improved identity preservation and video quality.
\subsection{Face Aware Prompt Enhancement}
% ------------- Prompt-Enhancement Module ----------------
% \paragraph{Prompt Enhancement.}
Given the raw text prompt $T$ and reference‐face image  $I^{ref}$ , we employ GPT-4o \cite{hurst2024gpt} to extract essential facial cues of $I^{ref}$ and inject them into the text without altering its original wording.  
The instruction supplied to the GPT-4o \cite{hurst2024gpt} 
is summarised below:

\begin{quote}\small
\textbf{Inputs}\\
1.~Original prompt $T$\\
2.~Image of one person’s face $I^{ref}$\\[2pt]
\textbf{Task}\\
Return \emph{one} revised caption according to the inputs that  
\begin{itemize}
  \item preserve the original caption verbatim;
  \item insert a short clause including only facial attributes of the image
        (approx.\ age, gender presentation,notable traits);
  \item omit clothing, accessories and background of the image;
  \item ensure the result reads as one natural sentence.
\end{itemize}

\end{quote}
The generated face aware prompt  $T_c$  ensures identity cues are presented (Figure. \ref{pipeline} $T_c$ including facial attributes like hair color and mustache), mitigating face omission issues in video generation.

\subsection{Prompt Aware Reference Image Enhancement}
To visually avoid semantic conflicts between the reference image $I^{ref}$ and the text prompt \(T_c\), we regenerate the reference image to include subject-specific attributes of the prompt \(T_c\). To acquire the prompt for regeneration, we parse the prompt \(T_c\) with GPT-4o~\cite{hurst2024gpt}. Since the new image is supposed to focus on subject-specific attributes of the prompt, such as the profession or clothes extracted from \(T_c\)(e.g., lifeguard and swimsuit in Figure \ref{pipeline}), the instruction for producing the image generation prompt is constructed as follows:

\begin{quote}\small
\textbf{Input}\\
Original prompt $T_c$\\[2pt]
\textbf{Task}\\
Return \emph{one} sentence that
\begin{itemize}
  \item preserves the subject’s identity and keeps the face fully visible (no occluding items);
  \item retains only profession/role attire, explicit actions, gender or hairstyle cues mentioned in the prompt;
  \item adds nothing not present in the description and focus on the attributes of the prompt subject;
  \item reads as natural third-person narration with no hashtags, camera directions, or meta language.
\end{itemize}
\end{quote}
The resulting reference-image prompt \(T_{\text{ref}}\), aligned with the video text prompt (e.g., in Figure. \ref{pipeline} "a person in a lifeguard uniform" to match the required appearance), is fed to the identity-preserving image generator~\cite{mao2025ace++} to produce the prompt-aware reference image \(I^{\text{ref}}_c\). This image, containing both the subject's identity and the prompt-specified attributes, then serves as the input to the video generator. Specifically, the revised image is extended from a single face crop to one that includes the prompt-specified appearance.

\subsection{ID-Aware Spatiotemporal Guidance 
Enhancement}
Existing video diffusion models rely on classifier-free guidance (CFG) \cite{ho2022classifier}, which compares conditional and unconditional text but cannot explicitly preserve identity or spatiotemporal quality, leading to identity drift and sub-optimal videos.  
We thus introduce ID-aware spatiotemporal guidance, which embeds identity preservation and quality objectives as gradients that steer the diffusion process.  
Given the noisy sample \(x_t\) at step \(t\) and the prompt \(T_c\), we compute the gradient of the log-probability of the reference identity \(I_c^{\mathrm{ref}}\) and imaginary high-quality video \(y_g\) to guide sampling:
\begin{align}
& \nabla_{x_t}\log p\!\bigl(  x_t\mid T_c,I_c^{\mathrm{ref}},y_g \bigr) \\
  &= \nabla_{x_t}\!\Bigl[ 
        \log p\!\bigl( x_t \mid T_c, I_c^{\mathrm{ref}} \bigr) \nonumber 
  \ - \log p\!\bigl( x_t \mid T_c,\varnothing,y_b \bigr)
     \Bigr] \\[4pt]
  &= \nabla_{x_t}\log p\!\bigl( x_t \mid T_c, I_c^{\mathrm{ref}} \bigr) \nonumber 
  \ - \nabla_{x_t}\log p\!\bigl( x_t \mid T_c,\varnothing,y_b \bigr)
\end{align}

This gradient encourages changes to $x_t$ that increase the likelihood of it belonging to the reference identity $I_c^{\mathrm{ref}}$ and high quality video $y_g$, by comparing the identity-conditioned density $p(x_t \mid T_c, I_c^{\mathrm{ref}})$ against the text-only and imaginary low quality label $y_b$ density $p(x_t \mid T_c, \varnothing,y_b)$. Following the classifier-free guidance principle, we treat our diffusion model as an implicit classifier that can estimate such probability gradients without an external classifier. The model’s score function (the gradient of the log-density with respect to $x_t$) is approximated by its denoising network $\epsilon_{\theta}$, which is trained to predict the noise residual. Therefore, the difference between the noise predicted by a high-quality video generation model under identity conditions and that predicted by a weaker model under identity-agnostic conditions can serve as an estimate of the aforementioned gradient.

 The ``weak" model's prediction $\epsilon_{\theta}^{b}(x_t \mid T_c, \varnothing)$ is obtained by removing identity inputs and skipping selected layers, producing a degraded, identity-agnostic network that outputs the low-quality label $y_b$.
The difference between the normal and weak predictions, $\epsilon_{\theta}(x_t \mid T_c, I_c^{\mathrm{ref}}) - \epsilon_{\theta}^{b}(x_t \mid T_c, \varnothing)$, explicitly captures identity-specific features and spatiotemporal details. Incorporating this difference as guidance reinforces identity preservation and video quality throughout denoising with minimal overhead.

Finally, we incorporate this identity guidance into the sampling step. We augment the conventional classifier-free guidance formulation by incorporating an additional identity preservation term. The final guidance signal used during sampling is:
\begin{align}
\tilde{\epsilon}_{\theta}(x_t) =\; 
&\epsilon_{\theta}\!\big(x_t \mid T_c, I_c^{\mathrm{ref}}\big)\;+\; 
W_c\,\Big[\epsilon_{\theta}\!\big(x_t \mid T_c, I_c^{\mathrm{ref}}\big)\;-\;\epsilon_{\theta}\!\big(x_t \mid \varnothing, I_c^{\mathrm{ref}}\big)\Big] \\[5pt]
&+\; W_i\,\Big[\epsilon_{\theta}\!\big(x_t \mid T_c, I_c^{\mathrm{ref}}\big)\;-\;\epsilon_{\theta}^b\!\big(x_t \mid T_c, \varnothing\big)\Big]~,
\end{align}
where $W_c$ and $W_i$ are weight hyperparameters for the original classifier-free guidance and ID-Aware spatiotemporal guidance. 

\begin{table*}
  \caption{Challenge Results of the Top Three Teams}
  \label{tab:team_results}
  \centering
  \small
  \begin{tabular}{c|c|c|c|c|c|c}
    \toprule
    Team Name & Face-Cur$\uparrow$ & Face-Arc$\uparrow$ & FID$\downarrow$ & ClipScore$\uparrow$ & User Study Score$\uparrow$ & Rank \\
    \midrule
    ghl (PKUVideo) - Ours & \textbf{0.492} & \textbf{0.473} & \textbf{170} & 27.8 & \textbf{1258} & 1 \\
    XuanYuan       & 0.467 & 0.441 & 214 & 28.0 & 1147.5 & 2 \\
    Wislab         & 0.285 & 0.269 & 208 & \textbf{28.6} & 594.5 & 3 \\
    \bottomrule
  \end{tabular}
\end{table*}

\begin{table*}
  \caption{Comparison of Different Methods}
  \label{tab:compare_all}
  \centering
  \small
  \begin{tabular}{c | cc | cc | ccc | c}
    \toprule
    \multirow{2}{*}{Methods}
      & \multicolumn{2}{c|}{Text Alignment} 
      & \multicolumn{2}{c|}{Identity Consistency} 
      & \multicolumn{3}{c|}{Video Quality} 
      & \multirow{2}{*}{OverallScore$\uparrow$} \\
    & CLIPScore$\uparrow$ & GMEScore$\uparrow$ 
     & CurScore$\uparrow$ & ArcScore$\uparrow$ 
     & Motion$\uparrow$ & Imaging$\uparrow$ & FID$\downarrow$ 
      & \\
    \midrule
    Hailuo  \cite{hailuo2024}  & \textbf{30.53} & 0.6277 & 0.0562 & 0.0452 & \textbf{0.9871} & \textbf{0.6793} & 249.63 & 0.4586 \\ \hline
    Phantom-14B \cite{liu2025phantom}  & 30.31 & \textbf{0.6399} & 0.2999 & 0.2847  & 0.9820 & 0.6364 & 251.15 & 0.5517 \\
    VACE-14B \cite{jiang2025vace} & 29.41 & 0.6217 & 0.3105 & 0.2983  & 0.9741 & 0.6294 & 217.32 & 0.5488 \\
    Ours   & 28.41 & 0.5990 & \textbf{0.4533} & \textbf{0.4358} & 0.9702 & 0.6441 & \textbf{184.44} & \textbf{0.5997} \\ 
    \bottomrule
  \end{tabular}
\end{table*}

\begin{table*}
  \caption{Ablation of Different Enhancements and MoE Strategy}
  \label{tab:ablation_enhance}
  \centering
  \small
  \begin{tabular}{c | c c | c c | c c c | c}
    \toprule
    \multirow{2}{*}{Methods}
      & \multicolumn{2}{c|}{Text Alignment} 
      & \multicolumn{2}{c|}{Identity Consistency} 
      & \multicolumn{3}{c|}{Video Quality} 
      & \multirow{2}{*}{OverallScore$\uparrow$} \\
    & CLIPScore$\uparrow$ & GMEScore$\uparrow$ 
      & CurScore$\uparrow$ & ArcScore$\uparrow$ 
      & Motion$\uparrow$ & Imaging$\uparrow$ & FID$\downarrow$ 
      & \\
    \midrule
    Baseline \cite{jiang2025vace} & 29.41 & \textbf{0.6217} & 0.3105 & 0.2983  & \textbf{0.9741} & 0.6294 & 217.32 & 0.5488 \\
    +PE & 28.89 & 0.6130 & 0.4040 & 0.3871 & 0.9734 & 0.6227 & 198.10 & 0.5815 \\
    +PE \& IE   & \textbf{30.24} & 0.6154 & 0.3510 & 0.3425 & 0.9737 & 0.6407 & 209.17 & 0.5655 \\ 
    +PE \& GE   & 28.41 & 0.5990 & \textbf{0.4533} & \textbf{0.4358} & 0.9702 & \textbf{0.6441} & \textbf{184.44} & \textbf{0.5997} \\ \hline
    
    MoE & -- & 0.6161 & \textbf{0.5176} & \textbf{0.5007}  & \textbf{0.9741} & \textbf{0.6606} & -- & \textbf{0.6337} \\
    \bottomrule
  \end{tabular}
\end{table*}

\begin{table*}
  \caption{Verification of Generalizability}
  \label{tab:ablation_gen}
  \centering
  \small
  \begin{tabular}{c | c c | c c | c c c | c}
    \toprule
    \multirow{2}{*}{Methods}
      & \multicolumn{2}{c|}{Text Alignment} 
      & \multicolumn{2}{c|}{Identity Consistency} 
      & \multicolumn{3}{c|}{Video Quality} 
      & \multirow{2}{*}{OverallScore$\uparrow$} \\
    & CLIPScore$\uparrow$ & GMEScore$\uparrow$ 
      & CurScore$\uparrow$ & ArcScore$\uparrow$ 
      & Motion$\uparrow$ & Imaging$\uparrow$ & FID$\downarrow$ 
      & \\
    \midrule
    Phantom-14B \cite{liu2025phantom}  & \textbf{30.62} & \textbf{0.6344} & 0.2611 & 0.2501  & 0.9785 & 0.6278 & \textbf{226.84} & 0.5335 \\
    Phantom-Enhance  & 30.41 & 0.6282 & \textbf{0.3232} & \textbf{0.3082} & \textbf{0.9839} & \textbf{0.6598} & 236.38 & \textbf{0.5613} \\
    
    \bottomrule
  \end{tabular}
\end{table*}

\begin{figure*}[t]          
  \centering
  \includegraphics[width=0.85\textwidth]{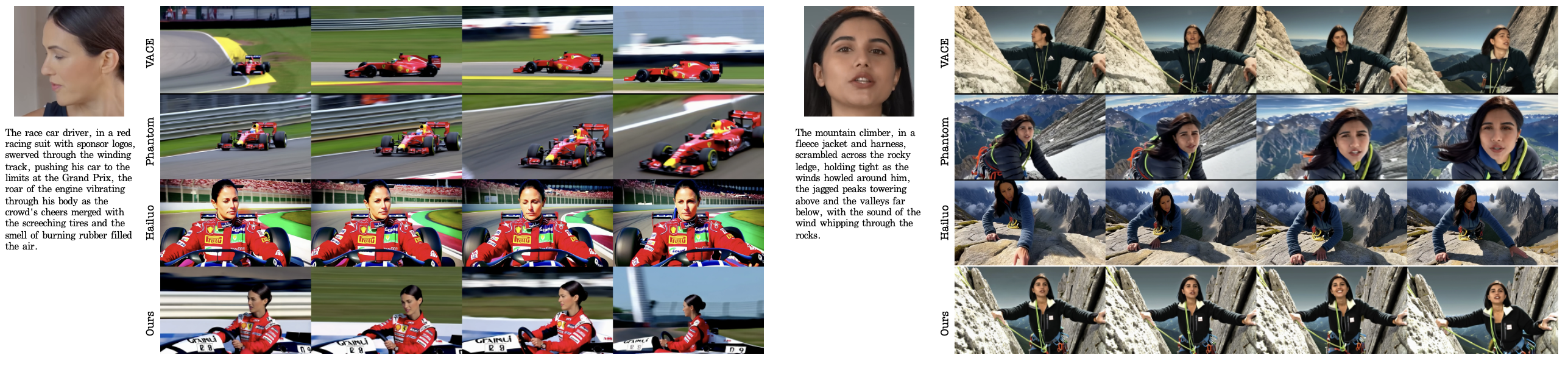}
  \caption{Visualization results of different methods}
  \label{visual}
\end{figure*}

\subsection{MoE Strategy}
\label{sec:moe}
Our proposed  Prompt Enhancement (PE), Image Enhancement (IE) and Guidance Enhancement (GE) methods each have distinct advantages, and there are many existing video generation models available for use.
To achieve the best challenge results, we adopted a Mixture of Experts(MoE) strategy to select the optimal output from different methods for each sample.

To determine which method performs best for a given sample, we need an overall score that evaluates the generated video from multiple dimensions \cite{yuan2025opens2v}. 

Specifically, We evaluate the video from three perspectives: \textbf{text alignment}, \textbf{identity consistency}, and \textbf{video quality}. (1) For \textbf{text alignment}, since previous methods using the CLIP model \cite{radford2021learning} faced issues such as a maximum token length of 77 (among the 50 prompts we sampled, 4 exceeded the length limit), we instead use the GME model \cite{zhang2024gme}, which is fine-tuned on Qwen2-VL \cite{wang2024qwen2}, to calculate the GMEScore. (2) For \textbf{identity consistency}, we compute the similarity between each generated frame and the reference image within the feature spaces of two facial recognition models, CurricularFace \cite{curricularface} and ArcFace \cite{arcface}, resulting in the CurScore and ArcScore metrics. (3) For \textbf{video quality}, we employ two metrics from VBench \cite{huang2024vbench}: Motion Smoothness, which leverages motion priors from a video frame interpolation model \cite{li2023amt} to assess the smoothness of generated motion, and Imaging Quality, which evaluates the presence of distortions such as over-exposure, noise, and blur in the generated frames. All metrics range from 0 to 1, with higher values indicating better performance.

Then we assign weights to each metric to calculate the overall score for a single video. The calculation formula is: $\text{Overall Score} = \sum_{i \in \mathcal{M}} w_i \cdot M_i$. In the formula, $\mathcal{M}$ is the set of evaluation metrics mentioned above, $w_i$ represents the weight of the corresponding evaluation metric. 
In this way, we can calculate the overall score for each video generated by different methods. For each sample, we evaluate six methods: the open-source models VACE \cite{jiang2025vace} and Phantom \cite{liu2025phantom}, the closed-source model Hailuo \cite{hailuo2024}, and three variants of our approach—PE, PE \& IE and PE \& GE. The video with the highest overall score among these six methods is selected as the final result for each sample.

\section{Experiments}
\subsection{Challenge Setup}
Identity-Preserving Video Generation (IPVG) task aims to generate videos from textual prompts while maintaining the consistency of the given reference identity throughout the text-to-video generation process. The challenge website is \url{https://hidream-ai.github.io/ipvg-challenge.github.io/}. 

\textbf{Test Dataset} 
The challenge’s test dataset contains 200 unseen person IDs. Each ID has portrait images and five textual prompts for video generation, totaling 1000 test pairs. 

\textbf{Evaluation Metric}
The challenge adopted the following evaluation metrics: Face-Cur and Face-Arc, which correspond to CurScore and ArcScore in Section \ref{sec:moe}, were used to measure identity preservation. FID \cite{FID} was used to assess feature differences in the face regions. Additionally, the CLIP score \cite{radford2021learning} was used to evaluate the similarity between the generated video and the text prompt, thereby determining text alignment.

\textbf{User Study}
The challenge adopted a user study for the top three teams based on the quantitative metrics. For each test sample, all pairwise combinations of the teams’ results were evaluated, resulting in 3,000 1v1 comparison pairs. In each head-to-head comparison, a win was awarded 1 point, a draw 0.5 point, and a loss 0 point.

\subsection{Our experimental setup}
\textbf{Dataset}
Since generating a test pair with the 14B video model takes a considerable amount of time, we sampled 50 unseen IDs and selected one unique prompt for each ID, ensuring that the prompts do not overlap. This resulted in an evaluation dataset with 50 samples, which is sufficient for this task, as validation datasets of a similar scale are also adopted in related works \cite{yuan2025opens2v, liu2025phantom}.

\textbf{Baselines}
We use VACE \cite{jiang2025vace} as the baseline and incorporate our proposed PE and GE strategies into this method as our approach. We compare our approach on the IPVG task with state-of-the-art open-source and closed-source models, including VACE, Phantom \cite{liu2025phantom}, and Hailuo \cite{hailuo2024}. Hailuo serves as a representative closed-source model because it achieves the best performance on this benchmark \cite{yuan2025opens2v}.
We combine the aforementioned metrics to comprehensively evaluate the performance of each method.

\subsection{Quantitative Results}
As shown in the Table \ref{tab:compare_all}, (1) our method significantly \textbf{outperforms other models on the two ID consistency metrics}, demonstrating that our enhancement strategies effectively improve the preservation of facial identity. (2) In terms of video quality, our method achieves the \textbf{best results on FID and Imaging Quality} among open-source methods, which is largely due to the implementation of GE that embeds quality objectives as gradients that steer the diffusion, thereby effectively enhancing the visual fidelity and overall quality of the generated videos. (3) Our method performs slightly worse than other methods in terms of text alignment. This may be because other methods are inferior in terms of facial preservation when aligning with the text, while our method takes both the text condition and the reference image condition into account during the video generation process.

\subsection{Qualitative Results}
In Figure \ref{visual}, we show visualization results from different methods, with each video represented by four evenly sampled frames. (1) In the first video, VACE and Phantom fail to generate a face, likely due to the “race car driver” identity, while our method produces a clear, reference-like face, outperforming Hailuo. (2) In the second video, the results of Phantom and Hailuo are not sufficiently consistent with the reference image. Our method generates videos with better facial preservation and higher quality than VACE. Overall, our TPIGE framework enables our method to generate videos more consistent with the reference image and more reasonable.

\subsection{Ablation Study}
\begin{figure}[t]          
  \centering
  \includegraphics[width=0.9\linewidth]{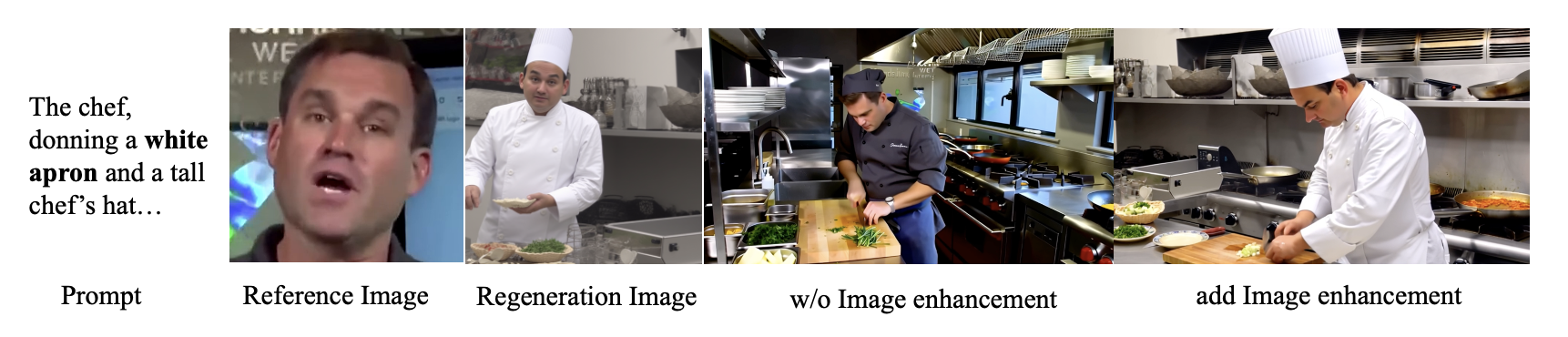}
  \caption{The impact of Image Enhancement}
  \label{image}
\end{figure}
\textbf{The Impact of Different Enhancements} 
We choose VACE \cite{jiang2025vace} as the baseline and incorporate various enhancement strategies on top of it. The results are reported in Table \ref{tab:ablation_enhance}:
(1) With the addition of PE, the identity consistency metric improves significantly. This is because PE injects facial feature information from the reference image into the original prompt, making the generated video more consistent with the reference image. While other metrics remain similar, the overall score increases noticeably.
(2) Building on PE, adding IE leads to improvements in text alignment and imaging quality metrics. This is because the regenerated reference image may have higher quality and incorporates more identity information from the original prompt, making the generated video better aligned with the prompt, as shown in Figure \ref{image}. However, this also results in a decrease in identity consistency. The fundamental reason is that the identity consistency metric is calculated based on similarity with the original reference image, while the generated video only receives the new reference image as input. Although the introduction of IE leads to a decrease in the overall score, it also improves text alignment and increases the diversity of the generated videos. Our MoE strategy also selects some videos generated by PE \& IE. 
(3) With the introduction of both PE and GE, the overall score reaches its best, and the metrics for identity consistency, imaging quality, and FID all achieve optimal performance. This indicates that, based on PE, GE further improves results by considering the reference image during video generation and is specifically designed to optimize video quality. Therefore, the combination of PE and GE leads to state-of-the-art performance.

\textbf{Effectiveness of the MoE Strategy}
We adopt the MoE strategy to select the video with the highest overall score for each sample from the results generated by different methods. We then compute the mean value of each metric (excluding CLIPScore and FID, as they are not involved in the calculation of the overall score) for the resulting set of videos. We found that videos generated by each method were selected, including those with the IE. This indicates each method has its own advantages. As shown in the last row of Table~\ref{tab:ablation_enhance}, almost all metrics reach the best performance, confirming the effectiveness of the MoE strategy for final challenge results.

\textbf{Generalizability of the TPIGE framework}
As shown in Table \ref{tab:ablation_gen}, we applied PE \& GE to Phantom \cite{liu2025phantom} on a sample dataset. The results show our enhancement strategies improve Phantom’s performance, confirming our approach’s generalizability.

\subsection{Challenge Results}

We finally adopted the MoE strategy mentioned in Section \ref{sec:moe} to complete our challenge submission because it achieved the best performance on our validation dataset, as demonstrated in Table \ref{tab:ablation_enhance}. Our submitted results secured first place in the IPVG challenge, which is illustrated in Table \ref{tab:team_results}. Specifically, our method achieved the highest scores on the quantitative metrics Face-Cur, Face-Arc, and FID, thanks to our enhanced face aware prompts and the use of reference image to guide the video generation process. This resulted in markedly superior identity preservation compared to other approaches. Furthermore, our method ranked first in the user study, with over 60\% of participants preferring our results, demonstrating that the combination of GE and the MoE strategy further boosted video quality. 
As previously discussed, our method integrates both text and image conditions, representing a balanced trade-off between the two.
Note our approach had slightly lower CLIP score, this might be because our method not only takes text conditions into account, but also incorporates image conditions, representing a balanced trade-off between the two.

\section{Conclusion}

We propose TPIGE, a training-free framework for identity-preserving text-to-video generation, mutually refine the quality of input reference images and prompts through \textit{Face Aware Prompt Enhancement} and \textit{Prompt Aware Reference Image Enhancement}, ensuring faithful facial identity retention even in complex scenarios. We introduce \textit{ID-Aware Spatiotemporal Guidance Enhancement} to jointly optimize identity preservation and video quality. Additionally, a MoE output selection strategy is employed to boost performance across diverse cases, enabling our approach to outperform existing methods according to automatic metrics and human study, culminating in a first-place finish in the ACM MM 2025 Identity-Preserving Video Generation Challenge.

\noindent\textbf{Acknowledgements.} This work was supported by the
grants from the National Natural Science Foundation of
China (62372014, 62525201, 62132001, 62432001, 62201014), Beijing Nova Program and Beijing Natural Science Foundation (4252040, L247006).

\bibliographystyle{ACM-Reference-Format}
\bibliography{sample-base}

\appendix

\end{document}